\def\year{2016}
\documentclass[letterpaper]{article}
\usepackage{aaai16}
\usepackage{times}
\usepackage{helvet}
\usepackage{courier}
\usepackage{url}
\usepackage{graphicx}
\usepackage{subcaption}
\usepackage{listings}
\usepackage{color}
\usepackage{booktabs}
\usepackage{subcaption}
\usepackage[usenames,dvipsnames]{xcolor}

\usepackage{mathtools}
\usepackage{amsfonts}
\usepackage{mathtools}
\newcommand{\defeq}{\vcentcolon=}
\newcommand{\EU}[2]{\mathrm{EU}_{#1} \left [ #2 \right ]}
\newcommand{\E}[2]{\mathop{\mathbb{E}}_{#1} \left [ #2 \right ]}

\lstdefinelanguage{JavaScript}{
  keywords={typeof, new, true, false, catch, function, return, null, catch, switch, var, if, in, while, do, else, case, break},
  keywordstyle=\bfseries,
  ndkeywords={class, export, boolean, throw, implements, import, this},
  ndkeywordstyle=\bfseries,
  identifierstyle=\color{black},
  sensitive=false,
  comment=[l]{//},
  morecomment=[s]{/*}{*/},
  commentstyle=\color{darkolivegreen}\ttfamily,
  stringstyle=\color{red}\ttfamily,
  morestring=[b]',
  morestring=[b]"
}

\lstdefinestyle{hilight}{
  moredelim=**[is][\bfseries\color{MidnightBlue}]{@}{@},
}

\definecolor{Green}{RGB}{50,200,50}

\begin{document}

 \title{Learning the Preferences of Ignorant, Inconsistent Agents}

\author{Owain Evans \\ University of Oxford
\And Andreas Stuhlm\"uller \\ Stanford University
\And Noah D. Goodman \\ Stanford University}

\maketitle
\begin{abstract}
\begin{quote}
  An important use of machine learning is to learn what people value. What posts or photos should a user be shown? Which jobs or activities would a person find rewarding? In each case, observations of people's past choices can inform our inferences about their likes and preferences. If we assume that choices are approximately optimal according to some utility function, we can treat preference inference as Bayesian inverse planning. That is, given a prior on utility functions and some observed choices, we invert an optimal decision-making process to infer a posterior distribution on utility functions. However, people often deviate from approximate optimality. They have false beliefs, their planning is sub-optimal, and their choices may be temporally inconsistent due to hyperbolic discounting and other biases. We demonstrate how to incorporate these deviations into algorithms for preference inference by constructing generative models of planning for agents who are subject to false beliefs and time inconsistency. We explore the inferences these models make about preferences, beliefs, and biases. We present a behavioral experiment in which human subjects perform preference inference given the same observations of choices as our model. Results show that human subjects (like our model) explain choices in terms of systematic deviations from optimal behavior and suggest that they take such deviations into account when inferring preferences.
\end{quote}
\end{abstract}
{\bf Keywords:} Bayesian learning, cognitive biases, preference inference

\section{Introduction}

The problem of learning a person's preferences from observations of their choices features prominently in economics \cite{hausman2011preference}, in cognitive science \cite{baker2011bayesian,ullman2009help}, and in applied machine learning \cite{jannach2010recommender,ermon2014learning}.
To name just one example, social networking sites use a person's past behavior to select what stories, advertisements, and potential contacts to display to them.
A promising approach to learning preferences from observed choices is to invert a model of rational choice based on sequential decision making given a real-valued utility function \cite{russell1995modern}.
This approach is known as {\em Inverse Reinforcement Learning} \cite{ng2000algorithms} in an RL setting and as {\em Bayesian Inverse Planning} \cite{baker2009action} in the setting of probabilistic generative models.

This kind of approach usually assumes that the agent makes optimal decisions up to ``random noise'' in action selection \cite{kim2014inverse,zheng2014robust}.
However, human deviations from optimality are more systematic.
They result from persistent false beliefs, sub-optimal planning, and from biases such as time inconsistency and framing effects \cite{kahneman1979prospect}.
If such deviations are modeled as unstructured errors, we risk mistaken preference inferences.
For instance, if an agent repeatedly fails to choose a preferred option due to a systematic bias, we might conclude that the option is not preferred after all.
Consider someone who smokes every day while wishing to quit and viewing their actions as regrettable. 
In this situation, a model that has good predictive performance might nonetheless fail to identify what this person values.

In this paper, we explicitly take into account structured deviations from optimality when inferring preferences.
We construct a model of sequential planning for agents with inaccurate beliefs and time-inconsistent biases (in the form of hyperbolic discounting).
We then do Bayesian inference over this model to jointly infer an agent's preferences, beliefs and biases from sequences of actions in a simple Gridworld-style domain.

To demonstrate that this algorithm supports accurate preference inferences, we first exhibit a few simple cases where our model licenses conclusions that differ from standard approaches, and argue that they are intuitively plausible.
We then test this intuition by asking impartial human subjects to make preference inferences given the same data as our algorithm.
This is based on the assumption that people have expertise in inferring the preferences of others when the domain is simple and familiar from everyday experience.
We find that our algorithm is able to make the same kinds of inferences as our human judges: variations in choice are explained as being due to systematic factors such as false beliefs and strong temptations, not unexplainable error.

The possibility of false beliefs and cognitive biases means that observing only a few actions often fails to identify a single set of preferences. We show that humans recognize this ambiguity and provide a range of distinct explanations for the observed actions. When preferences can't be identified uniquely, our model is still able to capture the range of explanations that humans offer. Moreover, by computing a Bayesian posterior over possible explanations, we can predict the plausibility of explanations for human subjects.

\section{Computational Framework}

Our goal is to infer an agent's preferences from observations of their choices in sequential decision problems.
The key question for this project is: how are our observations of behavior related to the agent's preferences? In more technical terms, what generative model \cite{tenenbaum2011grow} best describes the agent's approximate sequential planning given some utility function? Given such a model and a prior on utility functions, we could ``invert'' it (by performing full Bayesian inference) to compute a posterior on what the agent values.

The following section describes the class of models we explore in this paper. We first take an informal look at the specific deviations from optimality that our agent model includes. We then define the model formally and show our implementation as a {\em probabilistic program}, an approach that clarifies our assumptions and enables easy exploration of deviations from optimal planning.

\subsection{Deviations from optimality}

We consider two kinds of deviations from optimality: 

\subsubsection{False beliefs and uncertainty}

Agents can have false or inaccurate beliefs. We represent beliefs as probability distributions over states and model belief updates as Bayesian inference. Planning for such agents has been studied in work on POMDPs \cite{kaelbling1998planning}. Inferring the preferences of such agents was studied in recent work \cite{baker2014modeling,panella2014learning}. Here, we are primarily interested in the interaction of false beliefs with other kinds of sub-optimality.

\subsubsection{Temporal inconsistency}

Agents can be time-inconsistent (also called ``dynamically inconsistent''). Time-inconsistent agents make plans that they later abandon. This concept has been used to explain human behaviors such as procrastination, temptation and pre-commitment \cite{ainslie2001breakdown}, and has been studied extensively in psychology \cite{ainslie2001breakdown} and in economics \cite{laibson1997golden,o2000economics}.

A prominent formal model of human time inconsistency is the model of \emph{hyperbolic discounting} \cite{ainslie2001breakdown}. This model holds that the utility or reward of future outcomes is discounted relative to present outcomes according to a hyperbolic curve. For example, the discount for an outcome occurring at delay $d$ from the present might be modeled as a multiplicative factor $\frac{1}{1+d}$. The shape of the hyperbola means that the agent takes \$100 now over \$110 tomorrow, but would prefer to take \$110 after 31 days to \$100 after 30 days. The inconsistency shows when the 30th day comes around: now, the agent switches to preferring to take the \$100 immediately. 

This discounting model does not (on its own) determine how an agent plans sequentially. We consider two kinds of time-inconsistent agents. These agents differ in terms of whether they accurately model their future choices when they construct plans. First, a \emph{Sophisticated} agent has a fully accurate model of its own future decisions. Second, a \emph{Naive} agent models its future self as assigning the same (discounted) values to options as its present self. The Naive agent fails to accurately model its own time inconsistency.\footnote{The distinction and formal definition of Naive and Sophisticated agents is discussed in \citeauthor{o1999doing} (\citeyear{o1999doing}).}

\begin{figure}[t]
  \includegraphics[scale=0.24]{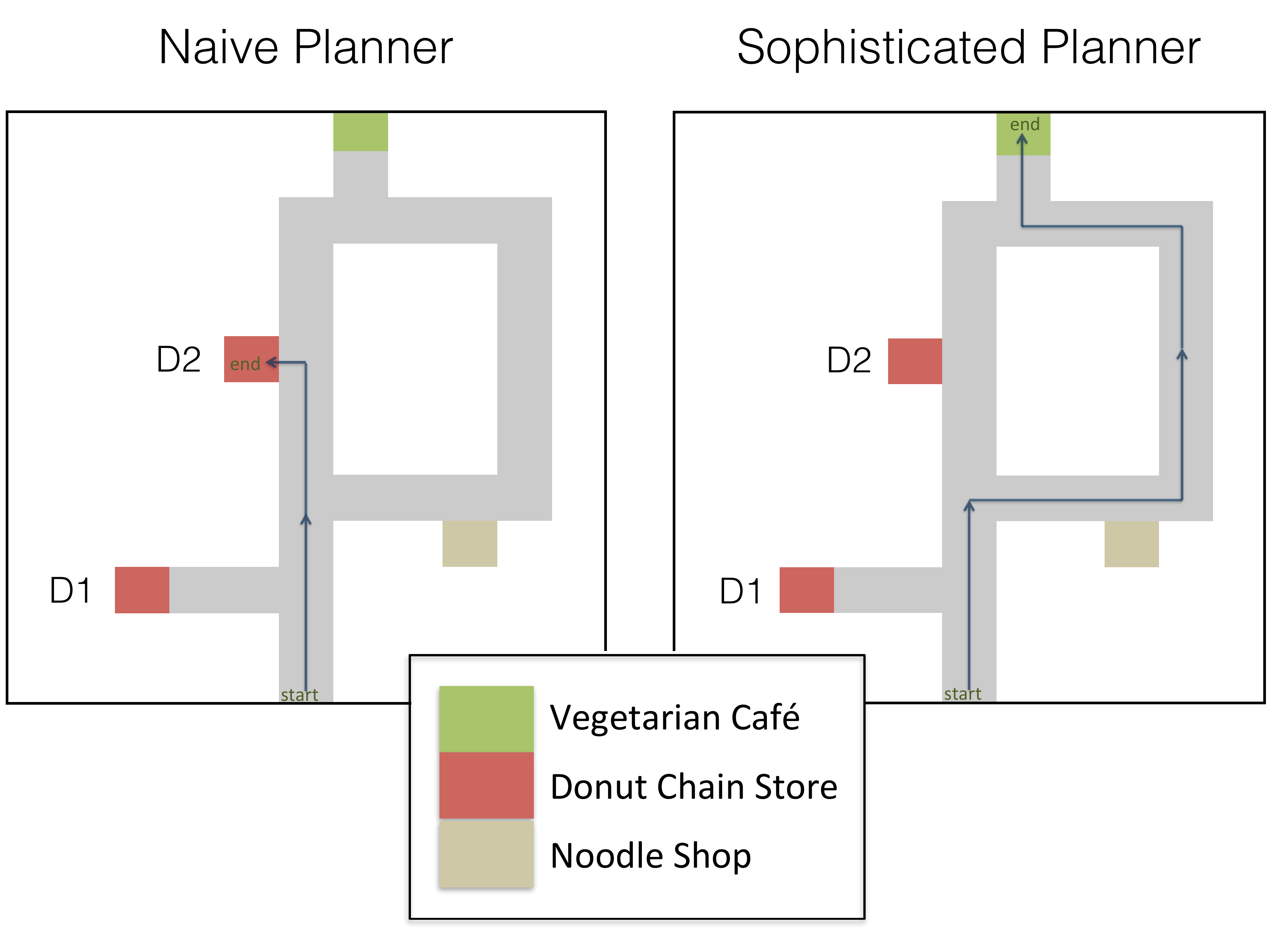}
  \caption{Agents with hyperbolic discounting exhibit different behaviors depending on whether they model their future discounting behavior in a manner that is (a) Naive (left) or (b) Sophisticated (right).}
  \label{fig:naive-sophisticated-grid}
\end{figure}

We illustrate Naive and Sophisticated agents with a decision problem that we later re-use in our experiments. The problem is a variant of Gridworld where an agent moves around the grid to find a place to eat (Figure \ref{fig:naive-sophisticated-grid}). 

In the left pane (Figure 1a), we see the path of an agent, Alice, who moves along the shortest path to the Vegetarian Cafe before going left and ending up eating at Donut Store D2.
This behavior is sub-optimal independent of whether her preference is for the Vegetarian Cafe or the Donut Store, but can be explained in terms of Naive time-inconsistent planning. From her starting point, Alice prefers to head for the Vegetarian Cafe (as it has a higher undiscounted utility than the Donut Store). She does not predict that when close to the Donut Store (D2), she will prefer to stop there due to hyperbolic discounting.

The right pane (Figure 1b) shows what Beth, a Sophisticated agent with similar preferences to Alice, would do in the same situation. Beth predicts that, if she took Alice's route, she would end up at the Donut Store D2. So she instead takes a longer route in order to avoid temptation. If the longer route wasn't available, Beth could not get to the Vegetarian Cafe without passing the Donut Store D2. In this case, Beth would either go directly to Donut Store D1, which is slightly closer than D2 to her starting point, or (if utility for the Vegetarian Cafe is sufficiently high) she would correctly predict that she will be able to resist the temptation.

\subsection{Formal model definition}

We first define an agent with full knowledge and no time inconsistency,\footnote{This is the kind of agent assumed in the  standard setup of an MDP \cite{russell1995modern}} and then generalize to agents that deviate from optimality.

We will refer to states $s \in S$, actions $a \in A$, a deterministic utility function $U\colon S \times A \rightarrow \mathbb{R}$, a stochastic action choice function $C\colon S \rightarrow A$, and a stochastic state transition function $T\colon S \times A \rightarrow S$. To refer to the probability that $C(s)$ returns $a$, we use $C(a; s)$.

\subsubsection{Optimal agent: full knowledge, no discounting}
Like all agents we consider, this agent chooses actions in proportion to exponentiated expected utility ({\em softmax}):
$$
C(a; s) \propto e^{\alpha \EU{s}{a}}
$$
The noise parameter $\alpha$ modulates between random choice ($\alpha=0$) and perfect maximization ($\alpha = \infty$).
Expected utility depends on both current and future utility:
$$
\EU{s}{a} = U(s, a) + \E{s', a'}{ \EU{s'}{a'}}
$$
with $s' \sim T(s, a)$ and $a' \sim C(s')$. Note that expected future utility recursively depends on $C$---that is, on what the agent assumes about how it will make future choices.

\subsubsection{Time-inconsistent agent}
Now the agent's choice and expected utility function are parameterized by a delay $d$, which together with a constant $k$ controls how much to discount future utility:
$$
C(a; s,d) \propto e^{\alpha \EU{s,d}{a}}
$$
$$
\EU{s,d}{a} = \frac{1}{1+kd} U(s, a) + \E{s', a'}{ \EU{s',d+1}{a'}}
$$
with $s' \sim T(s, a)$. For the Naive agent, $a' \sim C(s', d+1)$, whereas for the Sophisticated agent, $a' \sim C(s', 0)$. When we compute what the agent actually does in state $s$, we set $d$ to $0$. As a consequence, only the Sophisticated agent correctly predicts its future actions.\footnote{This foresight allows the Sophisticated agent to avoid tempting states when possible. If such states are unavoidable, the Sophisticated agent will choose inconsistently.} An implementation of the Naive agent as a probabilistic program is shown in Figure \ref{fig:agent-program}.

\subsubsection{Time-inconsistent agent with uncertainty}
We now relax the assumption that the agent knows the true world state. Instead, we use a distribution $p(s)$ to represent the agent's belief about which state holds. Using a likelihood function $p(o|s)$, the agent can update this belief:
$$
p(s|o) \propto p(s)p(o|s)
$$
The agent's choice and expected utility functions are now parameterized by the distribution $p(s)$ and the current observation $o$:
$$
C(a; p(s),o,d) \propto e^{\alpha \EU{p(s),o,d}{a}}
$$
To compute expected utility, we additionally take the expectation over states. Now $\EU{p(s),o,d}{a}$ is defined as:
$$
 \E{s \sim p(s|o)}{\frac{1}{1+kd} U(s, a) + \E{s', o', a'}{ \EU{p(s|o),o',d+1}{a'}}}
$$
with $s' \sim T(s, a)$, $o' \sim p(o|s')$ and $a' \sim C( p(s|o),o',d+1)$ (for the Naive agent) or $a' \sim C( p(s|o),o',0)$ (for the Sophisticated agent).

\begin{figure}[t]
  \begin{lstlisting}[style=hilight]
var @agent@ = function(state, delay){
 return Marginal(
  function(){
   var action = uniformDraw(actions)
   var eu = @expUtility@(state, action, delay)
   factor(alpha * eu)
   return action
  })
}

var @expUtility@ = function(state, action, delay){
 if (isFinal(state)){
  return 0
 } else {
  var u = 1/(1 + k*delay) * utility(state, action)
  return u + Expectation(function(){
    var nextState = transition(state, action)
    var nextAction = sample(@agent@(nextState, delay+1))
    return @expUtility@(nextState, nextAction, delay+1)  
  })
 }
}
\end{lstlisting}
\caption{We specify agents' decision-making processes as probabilistic programs. This makes it easy to encode arbitrary biases and decision-making constraints. When automated inference procedures invert such programs to infer utilities from choices, these constraints are automatically taken into account. Note the mutual recursion between \lstinline[style=hilight]{@agent@} and \lstinline[style=hilight]{@expUtility@}: the agent's reasoning about future expected utility includes a (potentially biased) model of its own decision-making.}
\label{fig:agent-program}
\end{figure}

\subsubsection{Inferring preferences}

We define a space of possible agents based on the dimensions described above (utility function $U$, prior $p(s)$, discount parameter $k$, noise parameter $\alpha$).  We additionally let $Y$ be a variable for the agent's \emph{type}, which fixes whether the agent discounts at all, and if so, whether the agent is Naive or Sophisticated. So, an agent is defined by a tuple $\theta \defeq (p(s),U,Y,k,\alpha)$, and we perform inference over this space given observed actions. The posterior joint distribution on agents conditioned on action sequence $a_{0:T}$ is:

\begin{equation}
  P(\theta | a_{0:T}) \propto P(a_{0:T}|\theta)P(\theta)
  \label{eqn:inference}
\end{equation}

The likelihood function $P(a_{0:T}|\theta)$ is given by the multi-step generalization of the choice function $C$ corresponding to $\theta$. 
For the prior $P(\theta)$, we use independent uniform priors on bounded intervals for each of the components.
In the following, ``the model'' refers to the generative process that involves a prior on agents and a likelihood for choices given an agent.

\subsection{Agents as probabilistic programs}

We implemented the model described above in the probabilistic programming language WebPPL \cite{dippl}. WebPPL provides automated inference over functional programs that involve recursion. This means that we can directly translate the recursions above into programs that represent an agent and the world simulation used for expected utility calculations. All of the agents above can be captured in a succinct functional program that can easily be extended to capture other kinds of sub-optimal planning. Figure \ref{fig:agent-program} shows a simplified example (including hyperbolic discounting but not uncertainty over state).

For the Bayesian inference corresponding to Equation \ref{eqn:inference} we use a discrete grid approximation for the continuous variables (i.e. for $U$, $p(s)$, $k$ and $\alpha$) and perform exact inference using enumeration with dynamic programming.

\subsection{Model inferences}
We now demonstrate that the model described above can infer preferences, false beliefs and time inconsistency jointly from simple action sequences similar to those that occur frequently in daily life. We later validate this intuition in our experiments, where we show that human subjects make inferences about the agent that are similar to those of our model. 

\begin{figure}[t]
  \centering
  \begin{minipage}[c]{.47\textwidth}
    \includegraphics[height=1.8in]{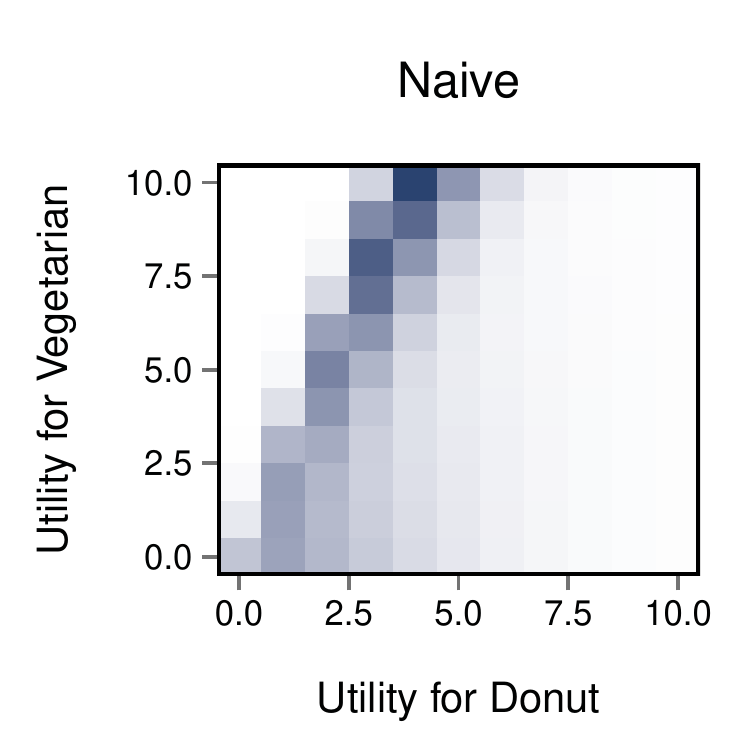}
    \includegraphics[height=1.8in]{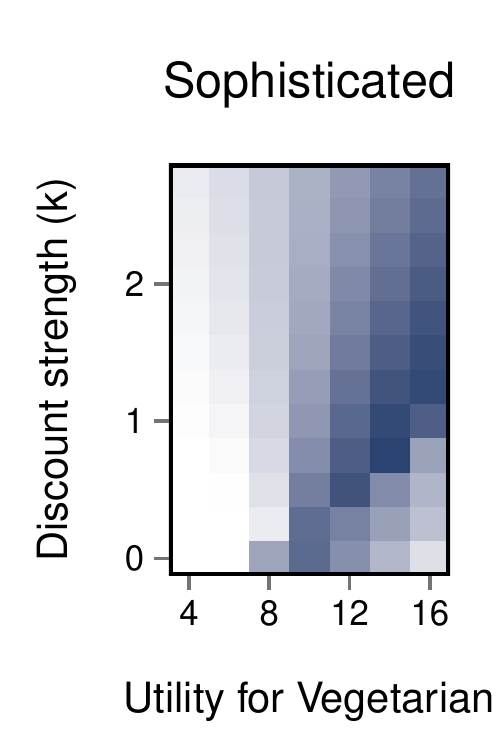}    
  \end{minipage}  
  \begin{minipage}[c]{0.2\textwidth}
\includegraphics[height=1.8in]{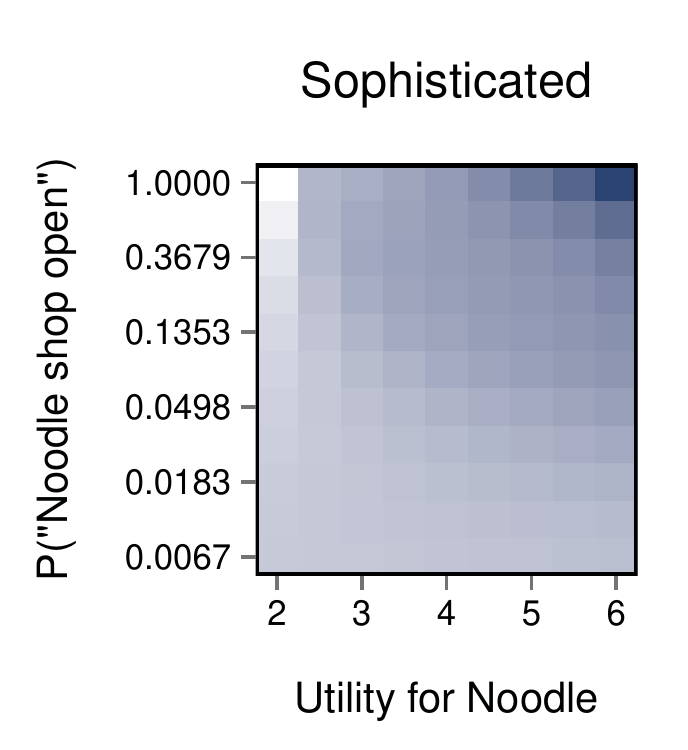}        
  \end{minipage}\hfill
  \begin{minipage}[c]{0.2\textwidth}
    \vskip 1em
    \caption{
      Given data corresponding to Figure \ref{fig:naive-sophisticated-grid}, the model infers a joint posterior distribution on preferences, beliefs and other agent properties (such as discount strength) that reveals relations between different possible inferences from the data. The darker a cell, the higher its posterior probability.
    } \label{fig:model-predictions}
  \end{minipage} 
\end{figure}

\subsubsection{Example 1: Inference with full knowledge}

We have previously seen how modeling agents as Naive and Sophisticated might predict the action sequences shown in Figures \ref{fig:naive-sophisticated-grid}a and \ref{fig:naive-sophisticated-grid}b respectively.
We now consider the inference problem. Given that these sequences are observed, what can be inferred about the agent?
We assume for now that the agent has accurate beliefs about the restaurants and that the two Donut Stores D1 and D2 are identical (with D1 closer to the starting point).\footnote{In Experiment 2, we allow the utilities for D1 and D2 to be different. See row 3 of Figure 6 and the ``Preference'' entry for Sophisticated in Figure 7.}
We model each restaurant as having an \emph{immediate} utility (received on arriving at the restaurant) and a \emph{delayed} utility (received one time-step after). This interacts with hyperbolic discounting, allowing the model to represent options that are especially ``tempting'' when they can be obtained with a short delay.

For the Naive episode (Figure \ref{fig:naive-sophisticated-grid}a) our model infers that either softmax noise is very high or that the agent is Naive (as explained for Alice above). If the agent is Naive, the utility of the Vegetarian Cafe must be higher than the Donut Store (otherwise, the agent wouldn't have attempted to go to the Cafe), but not too much higher (or the agent wouldn't give in to temptation, which it in fact does). This relationship is exhibited in Figure \ref{fig:model-predictions} (top left), which shows the model posterior for the utilities of the Donut Store and Vegetarian Cafe (holding fixed the other agent components $Y$, $k$, and $\alpha$).

\subsubsection{Example 2: Inference with uncertainty}

In realistic settings, people do not have full knowledge of all facts relevant to their choices. Moreover, an algorithm inferring preferences will itself be uncertain about the agent's uncertainty.
What can the model infer if it {\em doesn't} assume that the agent has full knowledge?
Consider the Sophisticated episode (Figure \ref{fig:naive-sophisticated-grid}b).
Suppose that the Noodle Shop is closed, and that the agent may or may not know about this.
This creates another possible inference, in addition to Sophisticated avoidance of temptation and high noise: The agent might prefer the Noodle Shop and might not know that it is closed.
This class of inferences is shown in Figure \ref{fig:model-predictions} (bottom):
When the agent has a strong prior belief that the shop is open, the observations are most plausible if the agent also assigns high utility to the Noodle Shop (since only then will the agent attempt to go there).
If the agent does not believe that the shop is open, the Noodle Shop's utility does not matter---the observations have the same plausibility either way.

In addition, the model can make inferences about the agent's discounting behavior (Figure \ref{fig:model-predictions} right): 
When utility for the Vegetarian Cafe is low, the model can't explain the data well regardless of discount rate $k$ (since, in this case, the agent would just go to the Donut Store directly).
The data is best explained when utility for the Vegetarian Cafe and discount rate are in balance---since, if the utility is very high relative to $k$, the agent could have gone directly to the Vegetarian Cafe, without danger of giving in to the Donut Store's temptation.

\begin{figure*}[t]
  \centering
  \includegraphics[scale=.48]{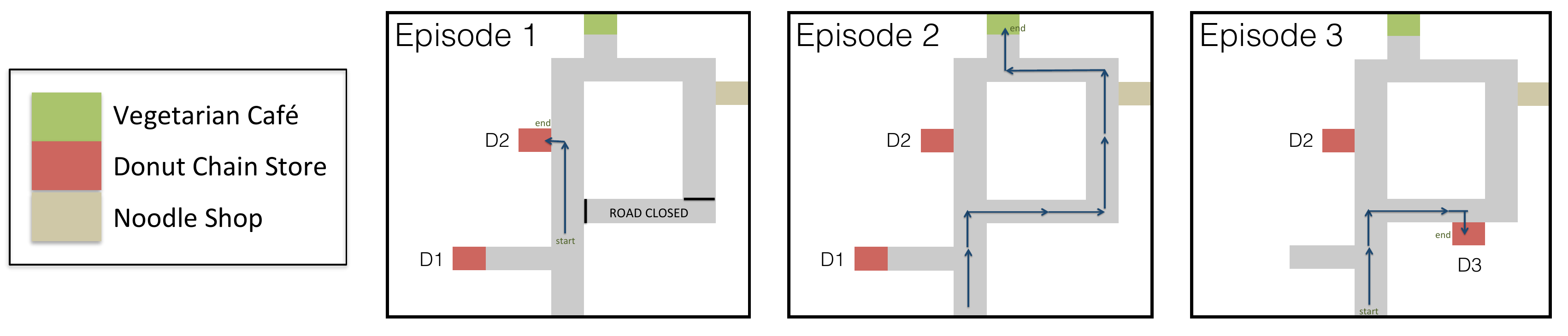}
  \caption{The observations in Experiment 3 show the Donut Chain Store being chosen twice and the Vegetarian Cafe once.}
  \vskip -.5em
  \label{fig:experiment-scenarios}
\end{figure*}

\subsubsection{Example 3: Inference from multiple episodes}
Hyperbolic discounting leads to choices that differ \emph{systematically} from those of a rational agent with identical preferences. A time-inconsistent agent might choose one restaurant more often than another, even if the latter restaurant provides more utility in total. Our model is able to perform this kind of inference. Figure \ref{fig:experiment-scenarios} shows the same agent choosing in three different episodes. While the agent chooses the Donut Store two out of three times, our model assigns posterior probability 0.59 (+/- 0.05 for 95\% CI) that the agent prefers the Vegetarian Cafe over the Donut Store.
As we decrease the prior probability of high softmax noise, this posterior increases beyond 0.59. By contrast, a model without time inconsistency infers a preference for the Donut Store, and has to explain Episode 2 in Figure \ref{fig:experiment-scenarios} in terms of noise, which leads to high-entropy predictions of future choices.

\section{Experiments with Human Subjects}

We have shown that, given short action sequences, our model can infer whether (and how) an agent is time-inconsistent while jointly inferring appropriate utilities.
We claim that this kind of inference is familiar from everyday life and hence intuitively plausible.
This section provides support for this claim by collecting data on the inferences of human subjects.
In our first two experiments, we ask subjects to explain the behavior in Figures \ref{fig:naive-sophisticated-grid}a and \ref{fig:naive-sophisticated-grid}b.
This probes not just their inferences about preferences, but also their inferences about biases and false beliefs that might have influenced the agent's choice. 

\subsection{Experiment 1: Inference with full knowledge}

Experiment 1 corresponds to Example 1 in the previous section (where the agent is assumed to have full knowledge). Two groups of subjects were shown Figures \ref{fig:naive-sophisticated-grid}a and \ref{fig:naive-sophisticated-grid}b, having already seen two prior episodes showing evidence of a preference for the Vegetarian Cafe over the other restaurants. People were then asked to judge the plausibility of different explanations of the agent's behavior in each episode.\footnote{In a pilot study, we showed subjects the same stimuli and had them write free-form explanations. In Experiment 1, subjects had to judge four of the explanations that occurred most frequently in this pilot.}

\begin{figure}
  \centering
  \vskip -.5em
  \includegraphics[height=2.5in]{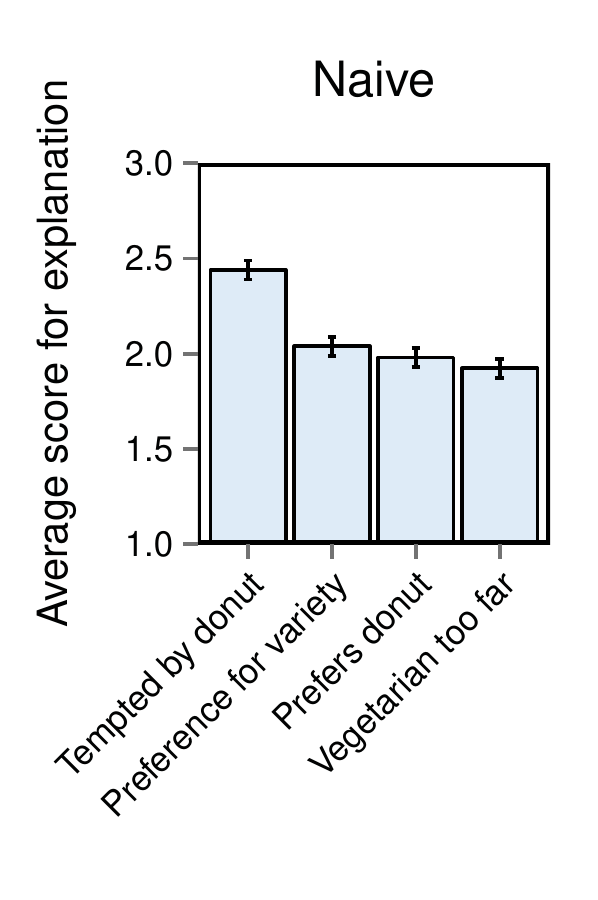}
  \hskip -1em
  \includegraphics[height=2.5in]{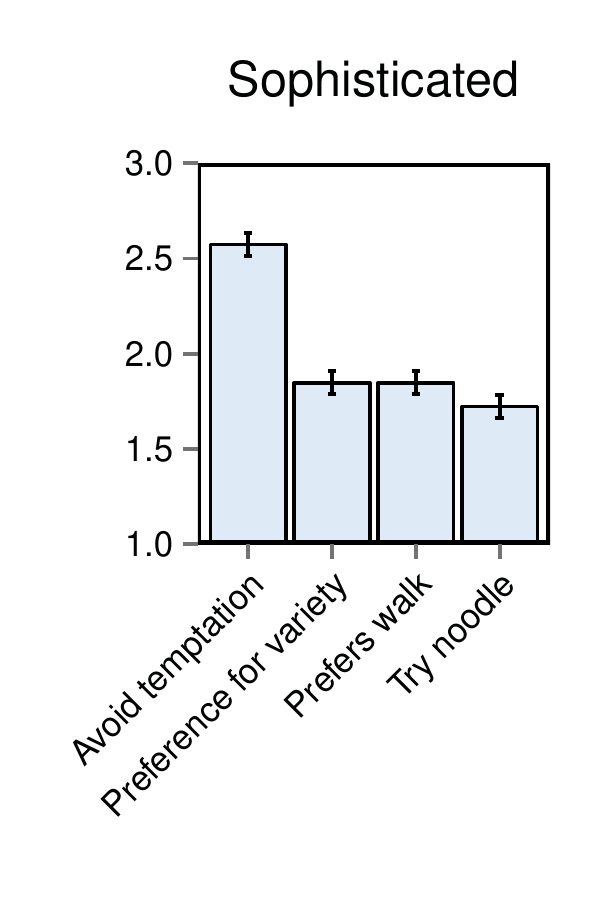}
  \vskip -1em
  \caption{Explanations in Experiment 1 for the agent's behavior in Figure 1a (Naive) and 1b (Sophisticated). Subjects (n=120) knew that the agent has accurate knowledge, and saw prior episodes providing evidence of a preference for the Vegetarian Cafe. Subjects selected scores in $\{1, 2, 3\}$.}
  \label{fig:exp12_explain}
\end{figure}

Results are shown in Figure \ref{fig:exp12_explain}. In both Naive (Figure 1a) and Sophisticated (1b) conditions, subjects gave the highest ratings to explanations involving giving in to temptation (Naive) or avoiding temptation (Sophisticated). Alternative explanations suggested that the agent wanted variety (having taking efficient routes to the Vegetarian Cafe in previous episodes) or that they acted purely based on a preference (for a long walk or for the Donut Store).

\subsection{Experiment 2: Inference with uncertainty}

Experiment 2 corresponds to Example 2 above. Subjects see one of the two episodes in Figure 1 (with Figure 1a modified so D1 and D2 can differ in utility and Figure 1b modified so the Noodle Shop is closed). There is no prior information about the agent's preferences, and the agent is not known to have accurate beliefs. We asked subjects to write explanations for the agent's behavior in the two episodes and coded these explanations into four categories.
Figure \ref{fig:property-table} specifies which formal agent properties correspond to which category.

\begin{figure*}[t]
  \centering		
  \begin{tabular}{ p{3.6cm} p{3.6cm} p{8.8cm} }
    \toprule
    Property & Formalization & Example explanation from our human subjects \\
    \midrule
    Agent doesn't know Donut Store D1 is open. & $p(D1=\mathrm{open}) < 0.15$ & ``He decided he wanted to go to the Donut Store for lunch. He did not know there was a closer location'' \\
    \midrule
    Agent falsely believes Noodle Shop is open. & $p(N=\mathrm{open}) > 0.85$ & ``He was heading towards the noodle shop first, but when he got there, it was closed, so he continued on the path and ended up settling for ... vegetarian cafe.'' \\
    \midrule
    Agent prefers D2 to D1. & $U(D2) > U(D1)$ & ``He might also enjoy the second donut shop more than the first'' \\
    \midrule
    Agent is Naive / Sophisticated. & $Y = \mathrm{Naive/Soph.}$ & ``He ... headed for the Vegetarian Cafe, but he had to pass by the Donut shop on his way.  The temptation was too much to fight, so he ended up going into the Donut Shop.'' \\
    \bottomrule		
\end{tabular}  		
  \caption{Map from properties invoked in human explanations to formalizations in the model. The left column describes the property. The center column shows how we formalized it in terms of the variables used to define an agent $\theta$. The right column gives an explanation (from our dataset of human subjects) that invokes this property.}		
  \label{fig:property-table}
\end{figure*}

\begin{figure}
  \centering
  \vskip -1em
  \includegraphics[height=2.4in]{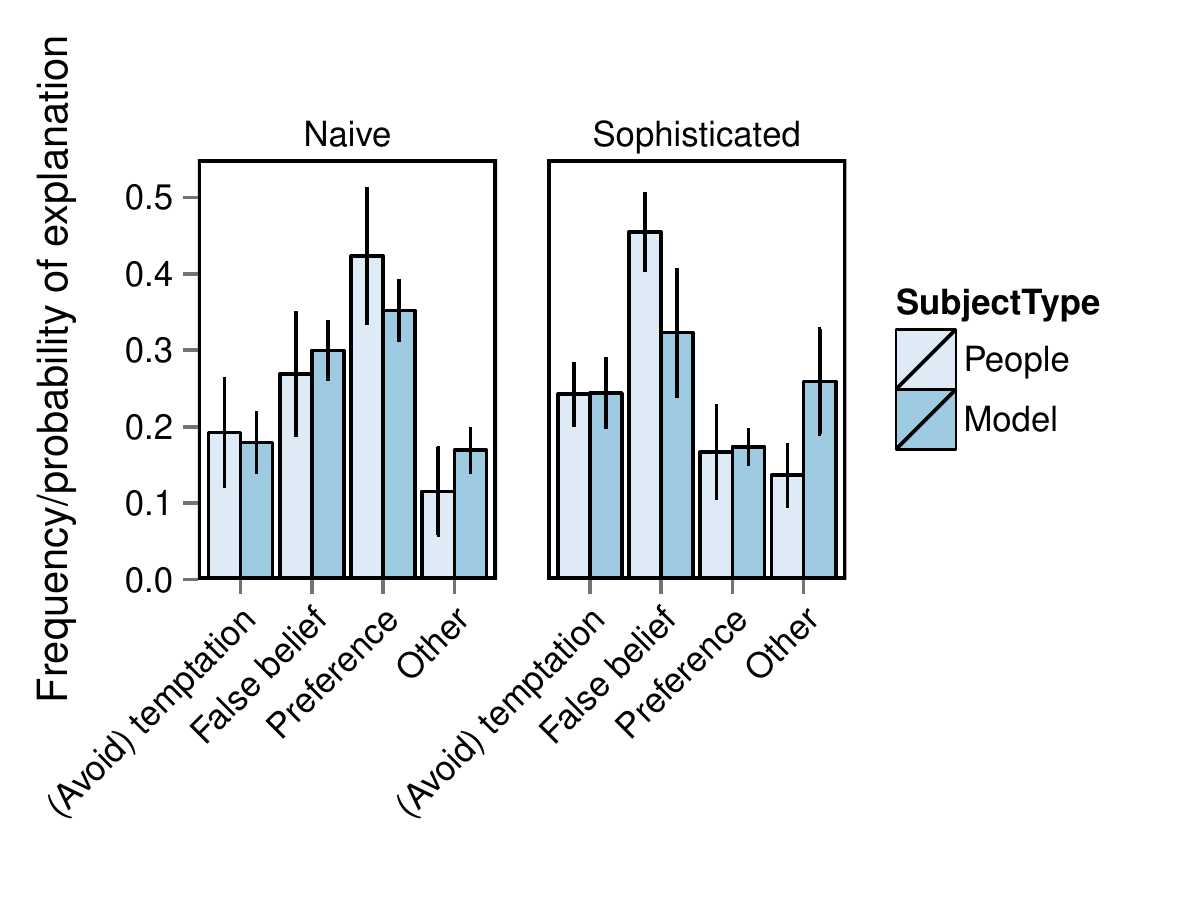}
  \vskip -1em
  \caption{Explanations in Experiment 2 for the agent's behavior in Figures 1a (Naive) and 1b (Sophisticated). Subjects did not know whether the agent has accurate knowledge, and did not see prior episodes. There were n=31 subjects (Naive) and n=40 subjects (Sophisticated).}
  \vskip -.5em  
  \label{fig:exp2}
\end{figure}

While not all explanations correspond to something the model can infer about the agent, the most common explanations map cleanly onto the agent properties $\theta$---few explanations provided by people fall into the ``Other'' category (Figure \ref{fig:exp2}). 
The model inferences in this figure show the marginal likelihood of the observed actions given the corresponding property of $\theta$, normalized across the four property types.
In both the Naive and the Sophisticated case, the model and people agree on what the three highest-scoring properties are.
Explanations involving false beliefs and preferences rate more highly than those involving time inconsistency.
This is because, even if we specify whether the agent is Naive/Sophisticated, the actions in Figure 1a/b are fairly unlikely---they require a narrow range of utility values, as illustrated in Figure \ref{fig:model-predictions} (top left), which favors more specific explanations.

\subsection{Experiment 3: Inference from multiple episodes}

Following Example 3 above, subjects (n=50) saw the episodes in Figure \ref{fig:experiment-scenarios} and inferred whether the agent prefers the Vegetarian Cafe or the Donut Store.
Like the model, the majority of subjects inferred that the agent prefers the Vegetarian Cafe.
Overall, 54\% (+/- 7 for 95\% CI) inferred a preference of Vegetarian Cafe over the Donut Store, compared to the 59\% posterior probability assigned by the model.
Episode 2 (in which the agent does not choose the Donut Store) is identical to the Sophisticated episode from Figure 1.
Experiments 1 and 2 showed that subjects explain this episode in terms of Sophisticated time-inconsistent planning.
Together with Experiment 3, this suggests that subjects use this inference about the agent's planning to infer the agent's undiscounted preferences, despite having seen the agent choose the Donut Store more frequently.

\subsection{Conclusion}

AI systems have the potential to improve our lives by helping us make choices that involve integrating vast amounts of information or that require us to make long and elaborate plans.
For instance, such systems can recommend and filter the information we see on social networks or music services and can construct intricate plans for travel or logistics.
For these systems to live up to their promise, we must be willing to delegate some of our choices to them---that is, we need such systems to \emph{reliably} act in accordance with our preferences and values.
It can be difficult to formally specify our preferences in complex domains; instead, it is desirable to have systems \emph{learn} our preferences, just as learning in other domains is frequently preferable to manual specification.

This learning requires us to build in assumptions about how our preferences relate to the observations the AI system receives. 
As a starting point, we can assume that our choices result from optimal rational planning given a latent utility function. However, as our experiments with human subjects show, this assumption doesn't match people's intuitions on the relation between preferences and behavior, and we find little support for the simplistic model where what is chosen most is inferred to be the most valued.
We exhibited more realistic models of human decision-making, which in turn supported more accurate preference inferences.
By approaching preference inference as probabilistic induction over a space of such models, we can maintain uncertainty about preferences and other agent properties when the observed actions are ambiguous.

This paper has only taken a first step in the direction we advocate. Two priorities for further work are applications to more realistic domains and the development of alternatives to using human preference inferences as a standard by which to evaluate algorithms. The goal for this emerging subfield of AI is to make systems better able to support humans even in domains where human values are complex and nuanced, and where human choices may be far from optimal.

\newpage

\section{Acknowledgments}
This work was supported by Future of Life Institute grant 2015-144846 (all authors) and by ONR grant N00014-13-1-0788 (NG).

This material is based on research sponsored by DARPA under agreement number FA8750-14-2-0009 (AS and NG). The U.S. Government is authorized to reproduce and distribute reprints for Governmental purposes notwithstanding any copyright notation thereon. The views and conclusions contained herein are those of the authors and should not be interpreted as necessarily representing the official policies or endorsements, either expressed or implied, of DARPA or the U.S. Government.

\nocite{*}
\bibliographystyle{aaai}
\bibliography{evans-stuhlmueller}

\end{document}